\documentclass[journal]{IEEEtran}
\usepackage{graphicx}
\usepackage{amsmath,amssymb} 
\usepackage{lineno}
\usepackage{color}
\usepackage{algorithmicx}
\usepackage{algpseudocode}
\usepackage{booktabs}
\usepackage{multirow}
\usepackage{epsfig}
\usepackage{subfigure}
\usepackage[ruled,vlined]{algorithm2e}
\usepackage{gensymb}

\begin{document}

\title{Maximum margin learning of t-SPNs for cell classification with filtered input}

% author information
\author{Haeyong~Kang,~\IEEEmembership{Student~Member,~IEEE,}, ~Chang~D.~Yoo,~\IEEEmembership{Senior~Member,~IEEE, ,~Yongcheon~Na,~\IEEEmembership{Member,~IEEE,}}% <-this % stops a space
\thanks{Haeyong Kang is with School of Electrical Engineering, Korea
	Advanced Institute of Science and Technology, 373-1 Guseong Dong,
	Yuseong Gu, Daejeon 305-701, South Korea.
	E-mail: ihaeyong@gmail.com.}% <-this % stops a space
\thanks{Chang D. Yoo is with School of Electrical Engineering, Korea
	Advanced Institute of Science and Technology, 373-1 Guseong Dong,
	Yuseong Gu, Daejeon 305-701, South Korea.
	E-mail: cdyoo@ee.kaist.ac.kr.}
\thanks{Yongcheon Na is with Interdisciplinary Program for Future Vehicles, Korea
	Advanced Institute of Science and Technology, 373-1 Guseong Dong,
	Yuseong Gu, Daejeon 305-701, South Korea.
	E-mail: ycna@kaist.ac.kr.}}% <-this % stops a space

 %in Future Vehicle from the Korea Advanced Institute of Science and Technology (KAIST) in 2015. 

\markboth{IEEE Selected Topics in Signal Processing Special Issue on Advanced Signal Processing in Microscopy and Cell Imaging}
{Shell \MakeLowercase{\textit{et al.}}: Bare Demo of IEEEtran.cls for Journals}
%

\maketitle

% This criterion in conjuction with L2-regularization seems to provide a good generalization and discriminability among the cells classes.
\begin{abstract}
An algorithm based on a deep probabilistic architecture referred to as a tree-structured sum-product network (t-SPN) is considered for cell classification. The t-SPN is constructed such that the unnormalized probability is represented as conditional probabilities of a subset of most similar cell classes. The constructed t-SPN architecture is learned by maximizing the margin, which is the difference in the conditional probability between the true and the most competitive false label. To enhance the generalization ability of the architecture, L2-regularization (REG) is considered along with the maximum margin (MM) criterion in the learning process. To highlight cell features, this paper investigates the effectiveness of two generic high-pass filters: ideal high-pass filtering and the Laplacian of Gaussian (LOG) filtering. On both HEp-2 and Feulgen benchmark datasets, the t-SPN architecture learned based on the max-margin criterion with regularization produced the highest accuracy rate  compared to other state-of-the-art algorithms that include convolutional neural network (CNN) based algorithms. The ideal high-pass filter was more effective on the HEp-2 dataset, which is based on immunofluorescence staining, while the LOG was more effective on the Feulgen dataset, which is based on Feulgen staining.
\end{abstract}


% \fi
\IEEEpeerreviewmaketitle

% -------------------------------------------------------------------------------------
%                      Chapter 1. Introduction 
% -------------------------------------------------------------------------------------
\section{Introduction}
\IEEEPARstart{S}{tudies} estimate prevalence of a broad group \footnote{more than 80 types} of autoimmune diseases \footnote{immune system producing antibodies against healthy cells} of over 3\%, and for effective treatment of these diseases, early and accurate diagnosis can significantly improve survival rates and quality of life. With the introduction of indirect immunofluorescence (IIF) and Feulgen staining methods, one effective means of diagnosis is to look for antinuclear antibodies. However, this task is time-consuming and labor-intensive, requiring highly specialized and experienced pathologists. The advancement of computer-aided diagnosis (CADx), founded on image processing and machine learning techniques, has transformed the diagnosis task into a cell classification problem to be resolved by computer software. In general, the incorporation of advanced machine learning techniques in CADx has the potential to revolutionize how histopathology is conducted.

Aside from any biological limitation associated with size and vagueness in biological patterns, cell classification accuracy is hindered by systematic limitations pertaining to the staining method: there is potentially a large variance in cell images captured by the reading system due to imprecise, nonuniform, and uncontrollable environmental conditions \cite{foggia2013benchmarking} related to slide preparation, conjugate specificity and the efficiency of the fluorescent microscope, thereby making cell classification a challenging problem. The issue can be further complicated by the photo-bleaching effect of a light source irradiating the cells over a short time with inconsistent illumination \cite{song1995photobleaching}.

% ========================= figure 1 ====================================
\begin{figure} [t]
	\centering
	\subfigure[Fine speckled.]{
		\label{fig:hep2_fs}
		\includegraphics[width=0.35\linewidth]{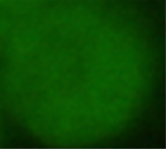}
		\label{fig:sl_fi}
	}\hspace{10mm}
	\subfigure[Homogeneous.]{
		\label{fig:hep2_hg}
		\includegraphics[width=0.35\linewidth]{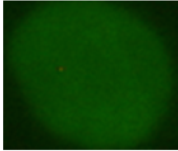}
		\label{fig:sl_ho}
	}\hfill \\
	\subfigure[Dystrophic mesothelial.]{
		\label{fig:fsc_dy}
		\includegraphics[width=0.35\linewidth]{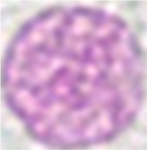}
		\label{fig:sl_dy}
	}\hspace{10mm}
	\subfigure[Normal mesothelial.]{
		\label{fig:fsc_me}
		\includegraphics[width=0.35\linewidth]{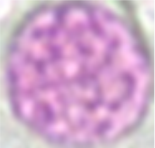}
		\label{fig:sl_me}
	}
	\caption{Samples of the two most confusing classes in each benchmark dataset: HEp-2 dataset (a) vs. (b); Feulgen stained dataset (c) vs. (d).}
	\label{fig:similar_looking}
\end{figure}

Concern as a consequence of biological and systematic limitations is the misclassification problem due to low contrast in the cell images: cell boundaries appear indistinct, leading to cell appearance and structure that are difficult to discriminate. Fig. \ref{fig:similar_looking} shows captured cell images of Human Epithelial Type 2 cells (HEp-2 cells) with dense, fine-speckled patterns (Fig. \ref{fig:sl_fi}) and with homogeneous patterns (Fig. \ref{fig:sl_ho}). These two images appear similar and are particularly difficult to distinguish. The Feulgen stained images (where chromatin takes a stoichiometric pink color while the cytoplasm is not colored) of dystrophic mesothelial (Fig. \ref{fig:sl_dy}) and normal mesothelial cell classes (Fig. \ref{fig:sl_me}) appear similar. They are difficult to discriminate against, even for a human expert.

To overcome the limitations mentioned above, past algorithms that have received considerable attention and success in improving the overall accuracy of cell classification have focused on feature extraction. Many approaches based on hand-crafted and learned features have been developed. Xiangfei \cite{kong2014hep} adopted a dictionary learning strategy for sparsely representing cell images. Nosaka \cite{nosaka2014hep}, the winner of the first \textit{HEp-2 Cell Classification Contest} \cite{foggia2013benchmarking}, proposed a novel descriptor, called rotation invariant co-occurrence among adjacent local binary patterns (LBP), which is invariant to local and global rotations of cell images and robust against constant changes in intensity. Kastaniotis \cite{theodorakopoulos2014hep} considered a discriminative sparse representation that combines an enhanced version of two well-known descriptors: scale-invariant feature transform (SIFT) and LBP. Features learned to use a deep architecture, particularly the convolutional neural networks (CNN), have been effective. Buyssens \cite{buyssens2013multiscale} proposed a multi-scale CNN (MCNN) that uses a multi-scaled input of Feulgen stained cell images. In all these methods, classification is based on maximizing individual class probability rather than improving discriminability between confusing cell classes.
	
%In this paper, for the cell classification, preprocessing  
Based on existing feature \cite{coates2011analysis} previously proposed, this paper proposes a deep probabilistic architecture referred to as the tree-structured sum-product networks (t-SPN) for cell classification. Measures are taken at the design and learning stages of the t-SPN to achieve the highest cell classification accuracy. A t-SPN is constructed based on sub-SPNs of the most confusing cell classes. Here, the t-SPN structure allows cell classification to be conducted in multi-stages, such that the most discernible to the least discernible subsets of cell classes are classified when going from the root of its tree structure to the leaves. To further enhance its discriminating ability, the t-SPN structure is learned via the maximum margin (MM) criterion, and L2-regularization is considered to improve generalization. The paper also studies the effectiveness of two popular filtering methods in improving robustness against low contrast in captured cell images. Initial studies show that cell classification based on different forms of staining given differing cell classes requires specific filtering to highlight the various characteristics of the captured images. 
	
This paper is organized as follows. In Section \ref{sec:spns}, SPN is briefly reviewed, and the concept of sub-SPNs of confusing classes within the framework of t-SPNs is described. Section \ref{sec:Max_margin_learn} introduces max-margin (MM) learning with L2-regularization (REG). Section \ref{sec:dis_feat_ext_} reviews a feature extraction method previously proposed \cite{coates2011analysis}. It also reviews two generic high-pass filtering methods. Section \ref{sec:exp} describes experiments to compare the performance of t-SPN against other state-of-the-art algorithms and investigates the effectiveness of two filtering methods. Finally, Section \ref{sec:conc} concludes and summarizes the paper.

% -------------------------------------------------------------------------------------
%                      Chapter 2. Sum-product networks 
% -------------------------------------------------------------------------------------
\section{Sum-product networks}\label{sec:spns}
A graphical architecture referred to as SPN, which has shown great potential in image classification and possesses the property that allows partition function to be represented in a tractable manner, is briefly reviewed in Section \ref{sec:intro_SPNs}. This is followed by an introduction to a particular SPN architecture referred to as t-SPN that factors the joint probability of the inputs into conditional probabilities of subsets of classes that are difficult to distinguish. In Section \ref{sec:intro_sub_SPNs}, the proposed t-SPN allows incorporation of divide-and-conquer type of philosophy in the classification such that classification can be performed in multi-stages such that the most discernible to the least discernible subset of cell classes are classified. Section \ref{sec:SPNs_learning} discusses a learning paradigm referred to as large margin maximization that aligns with the classification task in hand.

% ========================= figure 2 ====================================
\begin{figure}[t]                                                 
	\centering
	\includegraphics[width=0.80\linewidth]{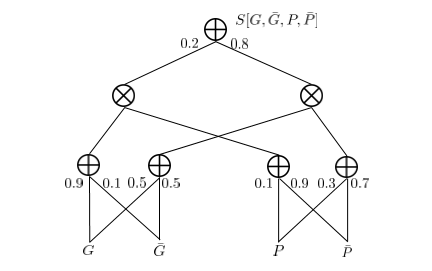}
	\caption{An example of a SPN S over Bernoulli distributed variables 'Granulars' $G$ and 'Pots' $P$ is shown such that $S[G,\bar{G},P,\bar{P}] = 0.2 \times (G \times 0.9 + \bar{G} \times 0.1)\times (P \times 0.1 + \bar{P} \times 0.9) + 0.8\times (G \times 0.5 + \bar{G} \times 0.5) \times (P \times 0.3 + \bar{P} \times 0.7)$.}
	\label{fig:naive_Bayes}
\end{figure} 

\subsection{Sum-product networks} \label{sec:intro_SPNs}
A Sum-Product Network (SPN) \cite{poon2011sum, kschischang2001factor, gens2012discriminative, gens2013learning, peharz2015theoretical, friesen2016sum, butz2020sum} is a rooted directed acyclic graph composed of sum and product nodes with input variables at the leaves. Fig. \ref{fig:naive_Bayes} shows an example of an SPN over two Bernoulli distributed binary input variables $G$ and $P$, which might be observed from a sample that belongs to either one of two classes: $G$ and $P$ are respectively Bernoulli distributed with probability 0.9 and 0.1 for one class and probability 0.5 and 0.3 for another class. when considering all possible outcomes, the probability of this distribution follows as $\bar{G}=1-G$ and $\bar{P}=1-P$. The sum and product nodes are utilized alternatingly, as shown in Fig. \ref{fig:naive_Bayes}. The edges entering the sum node are assigned non-negative weights, while the edges to the product nodes are not assigned any weights (or assigned a weight of 1). The sum node outputs the weighted sum of the inputs variables, while the product node outputs the product of its inputs. The weights assigned to the root sum node can be interpreted as the unnormalized class prior probabilities while the weights assigned to leaves can be interpreted as the unnormalized conditional likelihood of the input variables. It should be noted that all paths leading to a common input of the root sum node are associated to a common class. The unnormalized probability of evidence $S[G,\bar{G},P,\bar{P}]$ can be computed by performing a bottom-up evaluation of the valid SPN; the unnormalized conditional probability of each class is presented with $S[\boldsymbol{Y}|G,\bar{G},P,\bar{P}]$. As shown in Fig. \ref{fig:naive_Bayes},  as given $G=P=1$ and $\bar{G}=\bar{P}=0$, the unnormalized conditional probabilities are presented with $S[\boldsymbol{Y} = y_0|1,0,1,0 ] = 0.2 \times (0.9 \times 0.1) = 0.018$ and $S[\boldsymbol{Y} = y_1|1,0,1,0] = 0.8 \times (0.5 \times 0.3) = 0.12$ where output label can take either one of two labels such that $\boldsymbol{Y} \in \{y_0,y_1\}$. Additionally, The partition function $S[\boldsymbol{\rm{*}}] = S[1,1,1,1]$ can be obtained at the root as the input variables $G$, $\bar{G}$, $P$ and $\bar{P}$ are all set to $1$; all possible probabilities are considered as input of the SPN so that each class probability is presented with $S[\boldsymbol{Y}=y_0| 1,1,1,1] = 0.2$ and $S[\boldsymbol{Y}=y_1|1,1,1,1] = 0.8$ respectively.

% Sum-product Architecture for Confusing Cells classes 
\subsection{Sub-SPNs for confusing classes}\label{sec:intro_sub_SPNs}% ch. 
The philosophy of divide-and-conquer is incorporated into the SPN. Classification is conducted in multi-stages, such that the most discernible to the least discernible subsets of classes are classified. A number of SPNs referred to as sub-SPNs are constructed from subsets of similar cell classes, and from sub-SPNs that are similar, a larger sub-SPN can be constructed. This step can go on iteratively until only one sub-SPN includes all the cell classes left. This final SPN is referred to as t-SPN. Subsets of similar classes are determined based on a confusion matrix constructed based on a preliminary classification result using a plain SPN structure that considers classes indiscriminately.

\begin{figure}[t]
	\centering
	\subfigure[A SPN of two variables with the univariate distribution (the Bernoulli distribution), 'Granulars' $G$ and 'Pots' $P$.]{
		\label{fig:spns}
		\includegraphics[width=0.80\linewidth]{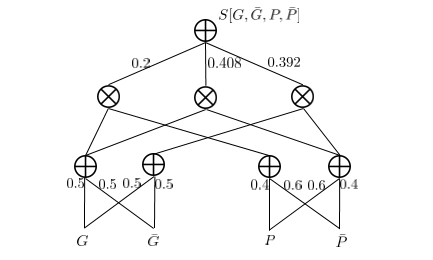}}
	\subfigure[A t-SPN of one variables with the univariate distribution (the Bernoulli distribution) 'Fibers' $F$ and a sub-SPN of two variables 'Granulars' $G$ and 'Pots' $P$.]{
		\label{fig:subspns}
		\includegraphics[width=0.80\linewidth]{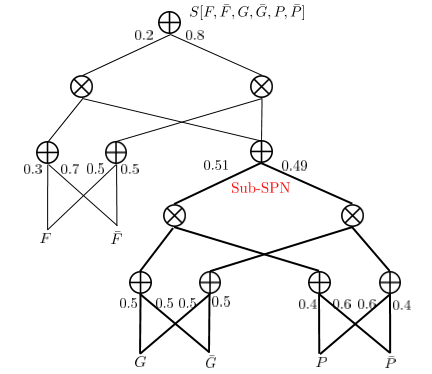}
	}
	\caption{A SPN and a t-SPN for a three classification task.}
	
	\label{fig:spn_arch}
\end{figure}

To understand the benefit of using t-SPN, two examples of SPNs are shown in Fig. \ref{fig:spn_arch}. In Fig. \ref{fig:spns}, all input variables are denoted by $S[G,\bar{G},P,\bar{P}]$. Each unnormalized probability of three classes is $S[\boldsymbol{Y} = y_0|1,0,1,0 ] = 0.2 \times (0.5 \times 0.4) = 0.04$, $S[\boldsymbol{Y} = y_1| 1,0,1,0] = 0.408 \times (0.5 \times 0.6) = 0.1224$, and $S[\boldsymbol{Y} = y_2|1,0,1,0] = 0.392 \times (0.5 \times 0.6) = 0.1176$. $S[\boldsymbol{Y} = y_0|\boldsymbol{*}] = 0.2$, $S[\boldsymbol{Y} = y_1|\boldsymbol{*}] = 0.408$ and $S[\boldsymbol{Y} = y_2|\boldsymbol{*}] = 0.392$ when all indicators of the variables are in a true state. The closest probabilistic difference between the two classes determines the most confusing classes. The confusing classes $S[\boldsymbol{Y} = y_1|\boldsymbol{*}]$ and $S[\boldsymbol{Y} = y_2|\boldsymbol{*}]$ design the sub-SPN in Fig. \ref{fig:subspns}. The classes are considered one class, so the weight of the root's edge is $0.8$. The variable 'Fibers' $F$ is added to focus on the probability of one class and the probabilities of the confusing classes. The classes with similar probabilities in Fig.\ref{fig:spns} explain the most confusing classes in Fig. \ref{fig:subspns}. The closer the probability is, the more likely a confusing result will occur. 

% ========================= figure 3 ====================================

This t-SPN is valid: all children of a sum node have the same scopes (complete) and under a product node, no variable appears negated in one child node and non-negated in another (consistent). The t-SPN is decomposable, which is somewhat more restrictive than consistency since the scopes of the children of each product node are disjoint.

\subsection{Learning t-SPN} \label{sec:SPNs_learning}
Learning of both SPNs and t-SPNs follows the hard gradient-based method for better accuracy \cite{gens2012discriminative}. There are two kinds of learning approaches in SPNs learning. One is the soft gradient-based method. Another is the hard gradient-based method. $S[\boldsymbol{y},\boldsymbol{h}|\boldsymbol{x}]$ is defined by the unnormalized conditional probability of a query $\boldsymbol{Y}=\boldsymbol{y}$, hidden variables $\boldsymbol{H}=\boldsymbol{h}$, and given $\boldsymbol{X}=\boldsymbol{x}$. In the process of discriminative training SPNs, it does not sum over states of given variables $\boldsymbol{X}$ that are treated as constants taking completeness and consistency into account. The gradient descent method $\boldsymbol{w}^{k+1}=\boldsymbol{w}^{k}+\alpha\frac{\partial}{\partial{\boldsymbol{w}}}\log{P(\boldsymbol{y}|\boldsymbol{x}^{\it{n}})} $ is used to update the parameters of SPNs and to maximize the conditional log-likelihood (CLL) where the normalized probability of $\boldsymbol{x}^n$ with the partition function is represented by $P(\boldsymbol{y}|\boldsymbol{x}^n)=S[\boldsymbol{y},\boldsymbol{h}|\boldsymbol{x}^n]/S[\boldsymbol{*},\boldsymbol{*}|\boldsymbol{x}^n]=S[\boldsymbol{y},\boldsymbol{h}|\boldsymbol{x}^n]/S[\boldsymbol{1},\boldsymbol{1}|\boldsymbol{x}^n]$, $\boldsymbol{w}$ is a weight of SPNs, $k$ is the iteration counter, $\alpha$ is the learning rate, $\boldsymbol{x}^{\it{n}}$ is a $n$th sample, and $\boldsymbol{y}$ is an output label of $\boldsymbol{x}^{\it{n}}$. Based on marginal inference, the soft gradient of CLL was derived as follows:
\begin{equation} % Eq. (1)
\begin{split}
\frac{\partial}{\partial{w}}\log{P(\boldsymbol{y}|\boldsymbol{x}^{\it{n}})} 
&=\frac{1}{S[\boldsymbol{y},\mathbf{1}|\boldsymbol{x}^{\it{n}}]}\frac{\partial{S[\boldsymbol{y},\boldsymbol{1}|\boldsymbol{x}^{\it{n}}]}}{\partial{w}}\\
&-\frac{1}{S[\boldsymbol{1},\boldsymbol{1}|\boldsymbol{x}^{\it{n}}]}\frac{\partial{S[\boldsymbol{1},\boldsymbol{1}|\boldsymbol{x}^{\it{n}}]}}{\partial{w}}.
\end{split}
\end{equation}
where the SPN with indicators set as $S[\boldsymbol{y},\boldsymbol{1}|\boldsymbol{x}]$ and $S[\boldsymbol{1},\boldsymbol{1}|\boldsymbol{x}]$ can be computed by bottom-up evaluations: $\boldsymbol{h} = \boldsymbol{1}$ presents that all $\boldsymbol{h}$ indicator functions is set to $1$, and $\boldsymbol{y} = \boldsymbol{1}$ shows the process of summing over the variable $\boldsymbol{y}$ (marginalization).      

The partial derivatives with respect to $\boldsymbol{w}$ can be calculated by backpropagation. For a sum node $i$, we get
$\frac{\partial{S}}{\partial{S}_{j}}\leftarrow\frac{\partial{S}}{\partial{S}_{j}}+w_{i,j}\frac{\partial{S}}{\partial{S}_{i}}$
and
$\frac{\partial{S}}{\partial{w_{i,j}}}\leftarrow
S_{j}\frac{\partial{S}}{\partial{S_{i}}}$
where $j$ is the child of parent (sum) node $i$ and $w_{i,j}$ is a weight of parent (sum) $i$. For a product node $i$, we obtain 
\begin{equation} % Eq. (2)
\begin{split}
\frac{\partial}{\partial{w}}\log{\widetilde{P}(\boldsymbol{y}|\boldsymbol{x}^{\it{n}})}
&=\frac{1}{M[\boldsymbol{y},\boldsymbol{1}|\boldsymbol{x}^{\it{n}}]}\frac{\partial{M[\boldsymbol{y},\boldsymbol{1}|\boldsymbol{x}^{\it{n}}]}}{\partial{w}}\\
&-\frac{1}{M[\boldsymbol{1},\boldsymbol{1}|\boldsymbol{x}^{\it{n}}]}\frac{\partial{M[\boldsymbol{1},\boldsymbol{1}|\boldsymbol{x}^{\it{n}}]}}{\partial{w}}.
\end{split}  
\end{equation}
The value of MPNs can be represented as $\prod_{w_{i}\in{W}}w_{i}^{c_{i}}$ where $W$ is a set of weights which appear in the branching path, and $c_{i}$ is the number of times $w_{i}$ appears in $W$. The partial derivative of MPN with respect to weight is $\frac{\partial{M}}{\partial{w_{i}}}=c_{i}\cdot{w_{i}^{c_{i}-1}}\prod_{w_{j}\in{W}\setminus\lbrace{w_{i}}\rbrace}w_{j}^{c_{j}} / \prod_{w_{j}\in{W}}w_{j}^{c_{j}} = c_{i}/w_{i}.$
% -------------------------------------------------------------------------------------
%                Chapter 3. Discriminative maximum margin learning 
% -------------------------------------------------------------------------------------
\section{Discriminative maximum margin learning of sum product networks} \label{sec:Max_margin_learn} 
In Section \ref{sec:Max_margin}, the maximum margin (MM) objective function for SPNs is introduced. Section \ref{sec:mm_learning} mentions MM-based learning of t-SPNs. 

\subsection{Maximum margin objective function} \label{sec:Max_margin} 
A margin-based-objective function is prepared for learning SPNs as well as t-SPNs. The objective function consists of a maximum margin (MM), a square hinge loss function, and a regularization term. The MM derives from the multi-class margin, the combinational margin of two classes out of the number of classes. The square hinge loss function is newly modified for both differentiability and the advantages of L2-SVM \cite{tang2013deep} in deep learnings. And L2-regularization which is a general method in machine learning is used further for SPNs learning.   

Using convex optimization, the MM is proposed to represent the multi-class margin \cite{guo2012maximum} to Bayesian networks (BN). The objective of the MM approach is to maximize the margin between the conditional probability of one true label and the maximum conditional probability of the other labels. The MM $d^{n}$ of SPNs is defined by using multi-class margin \cite{pernkopf2012maximum} of $\it{n}$th sample as follows: 
\begin{equation} % Eq. (3)
\begin{split}
d^{n}
=\min_{\boldsymbol{y} \neq \boldsymbol{y}\it{^{n}}}\frac{P(\boldsymbol{y}\it{^{n}}|\boldsymbol{{x}}^{\it{n}})}{P(\boldsymbol{y}|\boldsymbol{x}^{\it{n}})}
&=\frac{P[\boldsymbol{y}^{\it{n}},\boldsymbol{1}|\boldsymbol{x}^{\it{n}}]}{\max_{\boldsymbol{y} \neq \boldsymbol{y}\it{^{n}}}P[\boldsymbol{y},\boldsymbol{1}|\boldsymbol{x}^{\it{n}}]}\\
&=\frac{S[\boldsymbol{y}^{\it{n}},\boldsymbol{1}|\boldsymbol{x}^{\it{n}}]}{\max_{\boldsymbol{y} \neq \boldsymbol{y}\it{^{n}}}S[\boldsymbol{y},\boldsymbol{1}|\boldsymbol{x}^{\it{n}}]},
\label{eq:margin}
\end{split}
\end{equation}
where $\boldsymbol{x}^{n}$ is a $n$th sample, $\boldsymbol{y}$ is an output label of $\boldsymbol{x}\it{^n}$, $\boldsymbol{y}\it{^n}$ is a true label of $\boldsymbol{x}\it{^n}$, and $\boldsymbol{h}$ indicator functions are set to $\boldsymbol{1}$. If the margin $d^{n}$ is greater than one, the $n$th sample is correctly classified, or else the sample is misclassified.

This multi-class margin $\it{d^n}$ cannot be applied to the objective function directly due to  the non differentiability of the max function in Eq. \eqref{eq:margin}. When parameterized by some constant $\eta\geq1$, the max function can be approximated by the softmax function $\max_{x}f(x)\approx\log[\sum_{x}\exp{({\eta}f(x))}]^{\frac{1}{\eta}}$, where $f(x)$ is a nonnegative function \cite{sha2007comparison}. The approximated differentiable multi-class log margin $\mathcal{D^{\it{n}}}$ is redefined by the softmax function as below:
\begin{equation} % Eq. (4)
\begin{split}
\mathcal{D^{\it{n}}}
&=\log\it{d^n}\\
&\simeq{\log}S[\boldsymbol{y}^{\it{n}},\boldsymbol{1}|\boldsymbol{x}^{\it{n}}]-\frac{1}{\eta}\log\sum_{\boldsymbol{y} \neq \boldsymbol{y}\it{^{n}}}S[\boldsymbol{y},\boldsymbol{1}|\boldsymbol{x}^{\it{n}}]^{\eta}.
\end{split}
\label{eq:max-margin}
\end{equation}

% ========================= figure 4 ====================================
\begin{figure}
	\centering
	\includegraphics[width=0.85\linewidth]{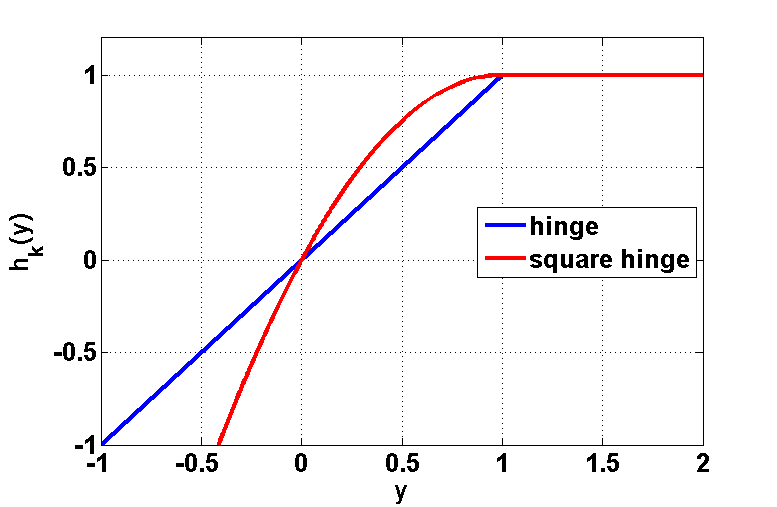}
	\caption{The hinge loss and its differentiable approximation for the square hinge loss. $y$ can be a multi-class margin ($\lambda\mathcal{D^{\it{n}}}$).}
	\label{fig:squarehinge}
\end{figure}

During the learning, it is desirable that the partial derivative of the objective with respect to weight is zero if $\mathcal{D^{\it{n}}}$ is more than one. Otherwise, the weight should be updated to the direction that $\mathcal{D^{\it{n}}}$ increases. The hinge loss function (Fig. \ref{fig:squarehinge}) is satisfied with those properties and follows $h_{k}(\lambda\mathcal{D^{\it{n}}})=\min(1,\lambda\mathcal{D^{\it{n}}})$
where ${\lambda}$ is the scaling parameter and $y^{n} = \lambda\mathcal{D^{\it{n}}}$. Since the hinge loss function is nondifferentiable at the point of $\lambda\mathcal{D^{\it{n}}}=1$, a differentiable square hinge loss function should be given below:
\begin{equation} % Eq. (5)
\begin{split}
h_{k}(\lambda\mathcal{D^{\it{n}}})=
\begin{cases}
1-(\lambda\mathcal{D^{\it{n}}}-1)^{2}, & \text{if}\: 			\lambda\mathcal{D^{\it{n}}}<1,\\
1, & \text{else}. %\: \lambda\mathcal{D^{\it{n}}}\geq 1.
\end{cases}
\end{split}
\label{eq:hinge_sq}
\end{equation}

The square hinge loss function imposes a bigger loss for points that violate the margin than the differentiable smooth hinge loss function does, proposed by Pernkopf {\it et al.} \cite{pernkopf2012maximum}. Our objective function of the square hinge loss function for maximizing the margin is defined by using Eq. \eqref{eq:hinge_sq} as follows:
\begin{equation} % Eq. (6)
H_{k}=\sum_{\it{n}}h_{k}(\lambda\mathcal{D^{\it{n}}})-\beta||\boldsymbol{w}||^{2}
\label{eq:obj_fcn}
\end{equation}
where $-\beta||\boldsymbol{w}||^{2}$ is L2-regularization term with the regularization parameter $\beta$ which proves this term in experiments and $\boldsymbol{w} = \log{S(\boldsymbol{y}|\boldsymbol{x}^{\it{n}})}$ is a weight vector of SPNs. As Eq. \eqref{eq:obj_fcn} is differentiable, the SGD method is applicable to optimize the objective function.

%-------------------------------------------------------------------------
\subsection{Maximum margin based learning of t-SPNs}\label{sec:mm_learning}
The weights are updated at training sample $n$ with SGD method to maximize the maximum margin given by Eq. \eqref{eq:obj_fcn} as follows $w^{k+1} = w^{k}+\alpha\frac{\partial}{\partial{w}}H_{k}$ where $k$ is the iteration counter, $\alpha$ is the learning rate. 
The gradient $\frac{\partial}{\partial{w}}h_{k}(\lambda\mathcal{D^{\it{n}}})$ is:
\begin{equation} % Eq. (7)
\begin{split}
\frac{\partial}{\partial{w}}H_{k}(\lambda\mathcal{D^{\it{n}}})=
\begin{cases}
-2\lambda(\lambda\mathcal{D^{\it{n}}}-1)\frac{\partial{\mathcal{D^{\it{n}}}}}{\partial{w}} -2\beta \boldsymbol{w}, & \text{if}\: \lambda\mathcal{D^{\it{n}}}<1,\\
0, & \text{else}. %\: \lambda\mathcal{D^{\it{n}}}\geq 1.
\end{cases}
\end{split}
\label{eq:hinge_sq_drv}
\end{equation}
where the partial derivative $-2\beta\boldsymbol{w}$ is added to the gradient when the L2-regularization is used to update the gradient, and the partial derivative of the multi-class log margin $\mathcal{D^{\it{n}}}$ based on the hard gradient follows:
\begin{equation} % Eq. (8)
\begin{split}
&\frac{{\partial}\widetilde{\mathcal{D}}^{n}}{\partial w} =\\
&\frac{1}{M[\boldsymbol{y}^{\it{n}},\boldsymbol{1}|\boldsymbol{x}^{\it{n}}]}\frac{\partial{M[\boldsymbol{y}^{\it{n}},\boldsymbol{1}|\boldsymbol{x}^{\it{n}}]}}{\partial{w}}\\
&-\frac{1}{\displaystyle\sum_{\boldsymbol{y} \neq \boldsymbol{y}\it{^{n}}}M[\boldsymbol{y},\boldsymbol{1}|\boldsymbol{x}^{\it{n}}]^{\eta}}\sum_{\boldsymbol{y} \neq \boldsymbol{y}\it{^{n}}}{\left(M[\boldsymbol{y},\boldsymbol{1}|\boldsymbol{x}^{\it{n}}]^{\eta -1}\frac{\partial {M[\boldsymbol{y},\boldsymbol{1}|\boldsymbol{x}^{\it{n}}}]}{\partial w}\right)}.
\end{split}
\label{eq:partial-max-margin}
\end{equation}
Then, the partial derivative of SPNs with respect to all lower layer weights can be updated by backpropagation with the hard inference (Section \ref{sec:SPNs_learning}).

% Structural benefits
In the t-SPN, this MM-based learning is approaching to maximize the margin among the classes, including a single class for the sub-SPN and learning to maximize the margin among the confusing classes in the sub-SPN. The way of learning and inferring a sub-SPN is equal to that of learning an SPN. This presents that the MM objective function for SPN learning derived in Section \ref{sec:Max_margin} is also used for updating the weights of the sub-SPN where the parameters of the sub-SPN depend on the number of confusing classes since the sub-SPN is regarded as the SPN of the small size. So, the MM function of Eq. \ref{eq:obj_fcn} can be easily applied to the sub-SPN. This sub-SPN learning method with MM approaches to minimizing MM-based loss among not only entire classes, including one class of confusing classes for the SPN but also the most confusing classes for the sub-SPN. 

% ========================= figure 5 ====================================
\begin{figure*}[t]
	\centering
	\includegraphics[width=1.0\linewidth]{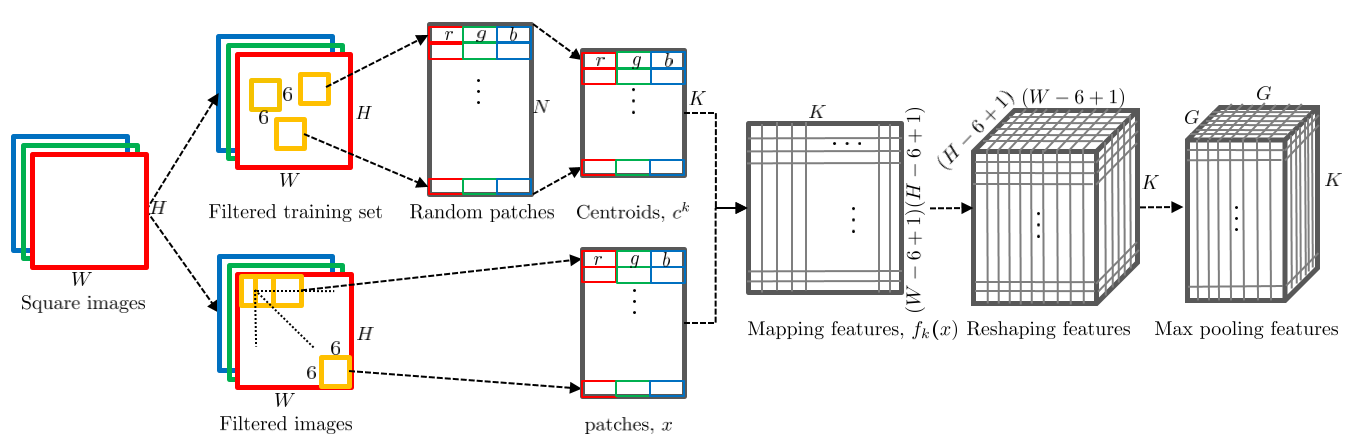}
	\caption{Feature extraction with filtering: $W$ is the width of the images, $H$ is the hight of the images, $N$ is the number of patches, $K$ is the number of centroids, $G\times G$ is the size of the max pooling matrix.}
	\label{fig:feat_ex}
\end{figure*}

% -------------------------------------------------------------------------------------
%                               Chapter 4. Discriminative Feature Extraction 
% -------------------------------------------------------------------------------------
\section{Discriminative Feature Extraction}\label{sec:dis_feat_ext_}
As mentioned earlier, this paper proposes a deep architecture referred to as the t-SPN for cell classification based on an existing feature previously proposed. This section will review the feature extraction method considered in \cite{coates2011analysis} for cell classification. The feature extraction takes high-pass filtered images as input, and this section will investigate the effectiveness of two generic high-pass filters.

\subsection{Filtering for discriminative features}\label{sec:dis_filter}
As mentioned earlier due to biological and systematic limitations, cell images captured by a reading system are generally of low contrast, and this is a major cause for misclassification. In the hopes of improving the sharpness, cell images are high-pass filtered to accentuate granular, dot, and skeleton patterns and textures. This paper investigates the effectiveness of two generic filtering methods, and preliminary results suggest that different staining would require a different filter.
 
\subsection{Feature extraction with filtering}\label{sec:dis_feat_ext}
The feature extraction follows the sequential approach proposed by Coates {\it et al.} \cite{coates2011analysis} with some modifications for cell classification. Fig.\ref{fig:feat_ex} shows how the features are extracted. (1) Training data and testing data sets are converted into square images by flipping the edges of the images such that it satisfies the feature extraction requirements. Only, the training set is augmented with rotations of 10$\degree$ steps. The square images are resized to the square images of $W \times H$ size to compare with other algorithms where $W$ is the width of the images and $H$ is the height of the images. (2) The square images are high-pass filtered to accentuate the texture information. (3) $400,000$ patches of size $6 \times 6$ are randomly extracted from the filtered training sets instead of striding. The patches are further processed by ZCA whitening \cite{bell1996edges}, which is commonly used in deep learnings \cite{krizhevsky2010factored}, followed by normalization. K-means clustering for 50 rounds generate $(6\times6\times3) \times K$ random patch centroids denoted as $\boldsymbol{c^{k}}$ in the figure. (4) Based on the centroids, the features of each image from the training and testing sets are computed: $6\times6$ size patches are extracted from the filtered image with stride 1 denoted as $\boldsymbol{x}$ in the figure. (5) The mapping features $z_{k}=||\boldsymbol{x}-\boldsymbol{c}^{(k)}||_{2}$ are computed for $0<k\leq{K}$. The mapping features are rectified at its mean: $f_{k}(\boldsymbol{x})=\max{\lbrace0,\mu{(z)}-z_{k}\rbrace}$ is computed where $\mu{(z)}$ is the mean of the $z_{k}$. Throughout this process, the mean of the $K$-dimensional feature is shifted to the origin. The values are then rectified into a $(W-6+1)\times(H-6+1)\times{K}$ feature matrix. Finally, max pooling is performed on the feature matrix, producing a $G\times{G}\times{K}$ feature matrix.  In the experiment, ${8}\times{8}\times{1600}$ sized features are used.

% -------------------------------------------------------------------------------------
%                               Chapter 5. Experiments 
% -------------------------------------------------------------------------------------
\section{Experiments} \label{sec:exp}% ch. 5
This section describes two benchmark datasets, the filtering approaches for discriminative feature extraction in Section \ref{sec:exp_data}, the experimental settings in Section \ref{sec:exp_setup}, the results in Section \ref{sec:exp_results} and \ref{sec:exp_reg}, and the comparison with other works in Section \ref{sec:exp_com}. 

\subsection{Datasets}\label{sec:exp_data}% ch. 5.1
% pre-processing data sets
The datasets are prepared for discriminatively training SPNs in a sequential manner. The augmented square HEp-2 cell images are resized to 96$\times$96, which is found in practice as a good default setting. The square Feulgen stained cell images are resized to 80$\times$80 for comparison with CNN \cite{buyssens2013multiscale}. Then, each dataset of the square cell images is processed in its own high-pass filtering (HPF) to focus on the texture information.\\

\begin{figure}[b]
	\centering
	\subfigure[Centromere.]{
		\label{fig:hep_2_ce}
		\includegraphics[width=0.29\linewidth]{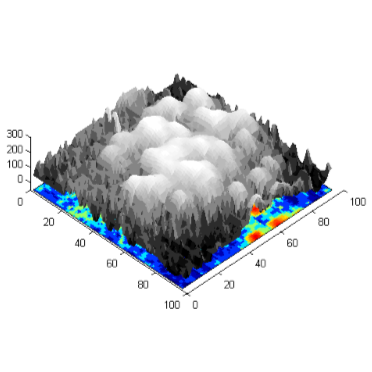}
	}
	\subfigure[Coarse speckled.]{
		\label{fig:hep_2_cs}
		\includegraphics[width=0.29\linewidth]{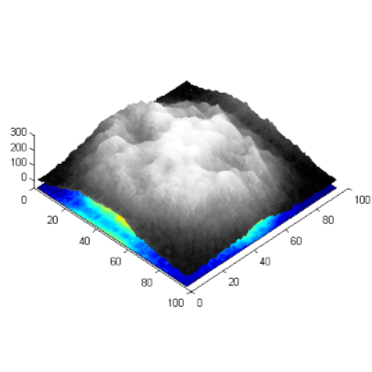}
	}
	\subfigure[Cytoplasmic.]{
		\label{fig:hep_2_cy}
		\includegraphics[width=0.29\linewidth]{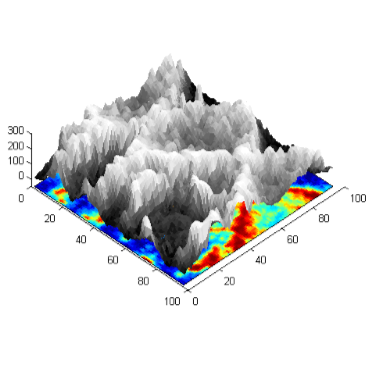}
	}\hfill
	\subfigure[Fine speckled.]{
		\label{fig:hep_2_fs}
		\includegraphics[width=0.29\linewidth]{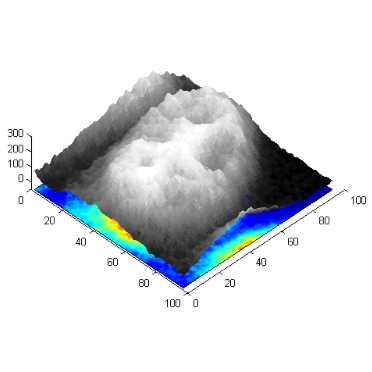}
	}
	\subfigure[Homogeneous.]{
		\label{fig:hep_2_ho}
		\includegraphics[width=0.29\linewidth]{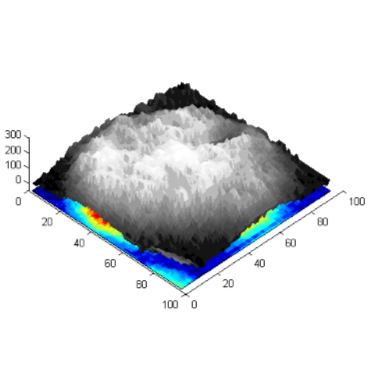}
	}
	\subfigure[Nucleolar.]{
		\label{fig:hep_2_nu}
		\includegraphics[width=0.29\linewidth]{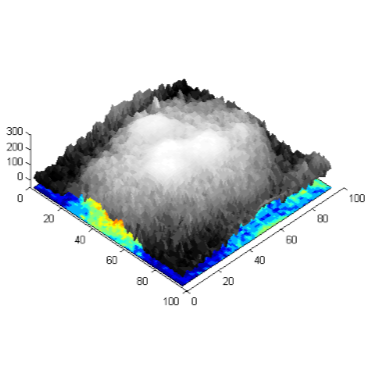}
	}
	\caption{Examples of square HEp-2 cell images shown in 3D of surf, Matlab.}
	\label{fig:hep-2-noHPF}
\end{figure}

\noindent {\bf HEp-2 cells}: HEp-2 \cite{nataro1987patterns, agata1994novel, perner2002mining, foggia2013benchmarking, sack2003computer, hiemann2009challenges, strandmark2012hep, foggia2014pattern, gao2016hep} dataset is one of the most widely used datasets in cell classification, as \textit{Indirect Immunofluorescence} (IIF) is increasingly becoming important in the diagnosis of several autoimmune diseases. The HEp-2 dataset$\footnote{http://mivia.unisa.it/datasets/biomedical-image-datasets/hep2-image-dataset/}$ is the official benchmark of the first \textit{HEp-2 Cell Classification Contest} \cite{foggia2013benchmarking}. 1,455 cells are selected from 28 images with a resolution of 1,388$\times$1,038 pixels and are split into 721 training and 734 test single-cell images. The HEp-2 dataset consists of six different classes with two intensity levels (intermediate and positive labels \cite{foggia2014pattern}). Fig. \ref{fig:hep-2-noHPF} shows the representative square HEp-2 cell image classes. The classification on SPNs is based on grayscale (green) images without intensity information.

% ========================= figure 6 ====================================

% ========================= figure 7 ====================================
\begin{figure}[b]
	\centering
	\subfigure[Centromere.]{
		\label{fig:hep_2_hpf_ce}
		\includegraphics[width=0.29\linewidth]{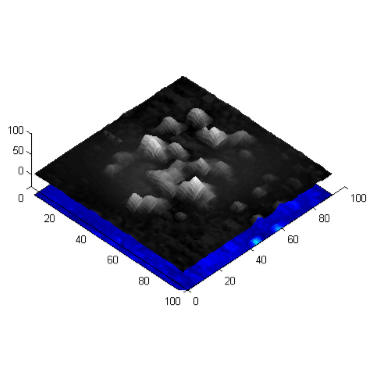}
	}
	\subfigure[Coarse speckled.]{
		\label{fig:hep_2_hpf_cs}
		\includegraphics[width=0.29\linewidth]{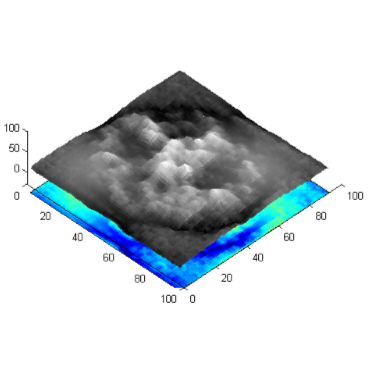}
	}
	\subfigure[Cytoplasmic.]{
		\label{fig:hep_2_hpf_cy}
		\includegraphics[width=0.29\linewidth]{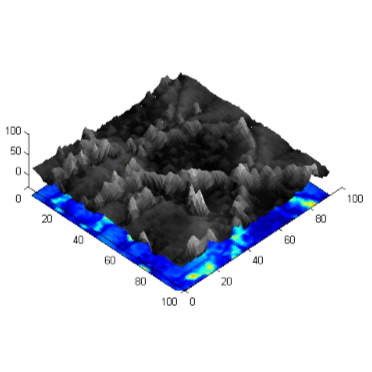}
	}\hfill
	\subfigure[Fine speckled.]{
		\label{fig:hep_2_hpf_fs}
		\includegraphics[width=0.29\linewidth]{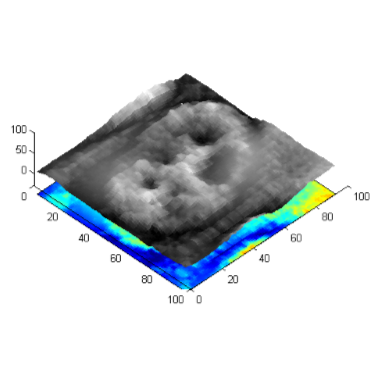}
	}
	\subfigure[Homogeneous.]{
		\label{fig:hep_2_hpf_ho}
		\includegraphics[width=0.29\linewidth]{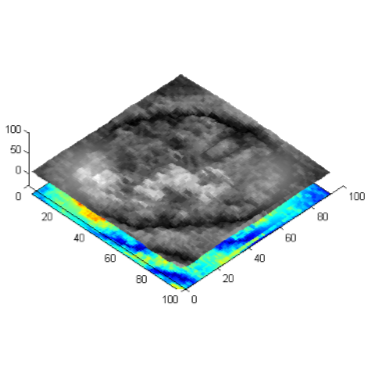}
	}
	\subfigure[Nucleolar.]{
		\label{fig:hep_2_hpf_nu}
		\includegraphics[width=0.29\linewidth]{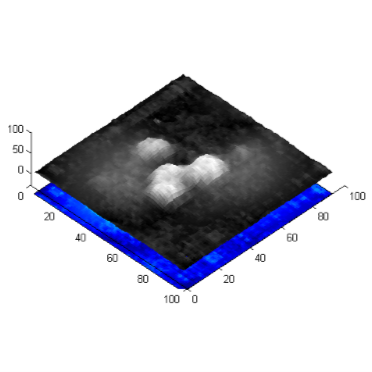}
	}
	\caption{Examples of HEp-2 cell images with high-pass filter shown in 3D of surf, Matlab.}
	\label{fig:hep-2-HPF}
\end{figure} 

% ========================= figure 8 ====================================
\begin{figure}[h]                                                 
	\centering
	\includegraphics[width=0.90\linewidth]{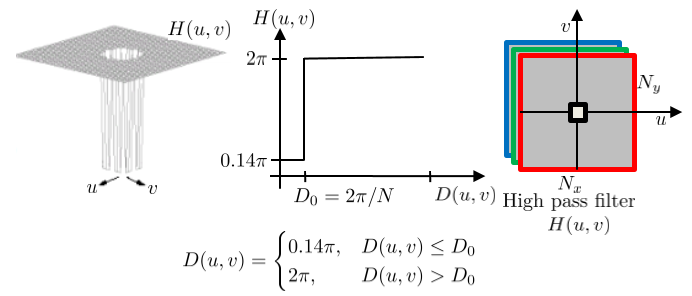}
	\caption{An ideal rectangular high-pass filter.}
	\label{fig:rec_hpf}
\end{figure} 

% Pre-processing (High pass filters)
An ideal rectangular high-pass filter similar to the ideal filter \cite[pp. 182-183]{gonzalez2002woods} can be used to sharpen the staining patterns of HEp-2 cell images $f(x,y)$ shown in Fig. \ref{fig:hep-2-noHPF}. The output of the high-pass filter in the frequency domain is computed by $G(u,v) = F(u,v)H(u,v)$, where $H(u,v)$ is the ideal rectangular high-pass filter as shown in Fig. \ref{fig:rec_hpf} and $F(u,v)$ is the discrete Fourier transform (DFT) of the cell images. In Fig. \ref{fig:rec_hpf}, $D(u,v)$ is the distance from point $(u,v)$ to the center of the frequency rectangle on each axis $(u, v)$, and $D_0 = 2\pi/N$ is the cut-off frequency, where $N = N_x = N_y$. $D(u,v)$ is $0.14\pi$ if $D(u,v) \leq D_0$, and $2\pi$ otherwise. The parameters of $N_x, N_y$ are the sizes of the cell images in the frequency domain. This ideal rectangular filter can be used as most of the cell information goes around the center of the frequency domain. That is, the ringing artifacts are so small that the ideal rectangular filter can be applied to the filtering of HEp-2 cell images. Therefore, the parameters of the filter are heuristically determined under the assumption that the features extracted by computers are also discriminative if the staining patterns can be clearly seen by the naked eye. The inverse DFT of $G(u,v)$ is the filtered cell images $g(x,y)$ as shown in Fig. \ref{fig:hep-2-HPF}. 

% ========================= figure 9 ====================================
\begin{figure}[b]
	\centering
	\subfigure[Abnormal mesothelials.]{
		\label{fig:fsc-nonLoG-AM}
		\includegraphics[width=0.29\linewidth]{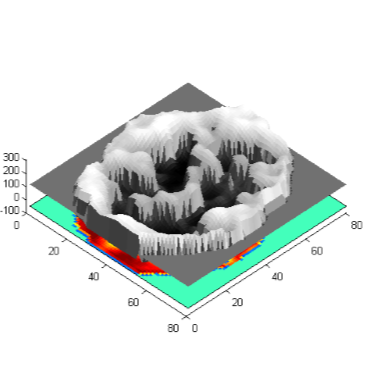}
	}
	\subfigure[Dystrophic mesothelials.]{
		\label{fig:fsc-nonLoG-DM}
		\includegraphics[width=0.29\linewidth]{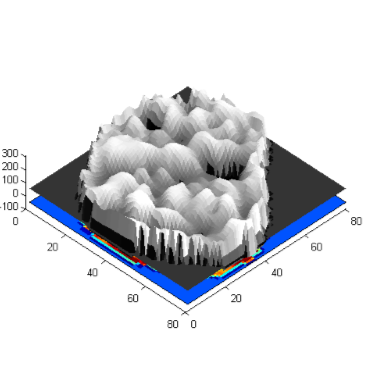}
	}
	\subfigure[Lymphocyte.]{
		\label{fig:fsc-nonLoG-Ly}
		\includegraphics[width=0.29\linewidth]{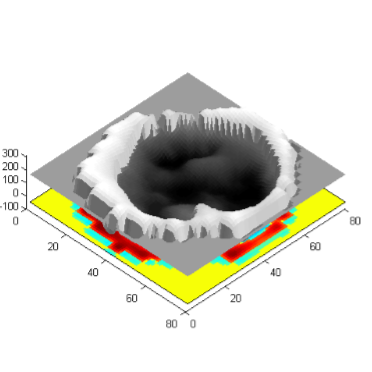}
	}\hfill
	\subfigure[Macrophages.]{
		\label{fig:fsc-nonLoG-Ma}
		\includegraphics[width=0.29\linewidth]{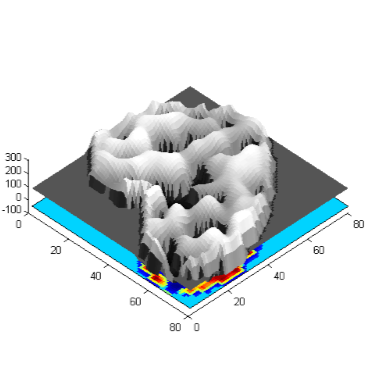}
	}
	\subfigure[Normal mesothelials]{
		\label{fig:fsc-nonLoG-NM}
		\includegraphics[width=0.29\linewidth]{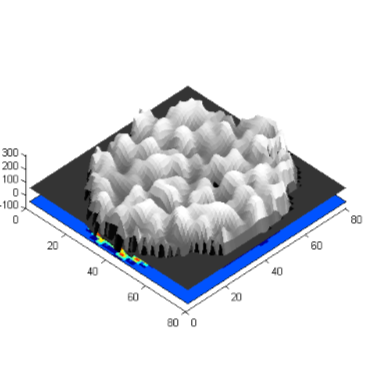}
	}
	\subfigure[Polynuclear.]{
		\label{fig:fsc-nonLoG-Po}
		\includegraphics[width=0.29\linewidth]{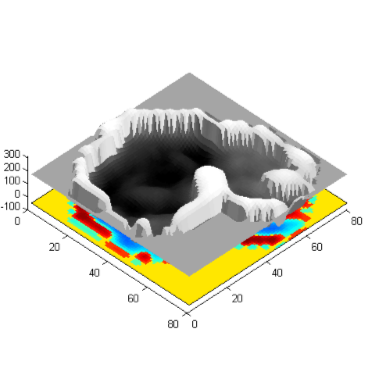}
	}
	\caption{Examples of Feulgen stained cell images shown in 3D of surf, Matlab.}
	\label{fig:fsc-noLoG}
\end{figure} 

The results of the high-pass filtering images visualized in 3D is shown in Fig. \ref{fig:hep-2-HPF}. Using this high-pass filter produces discriminative features for cell classification. The centromere cell image as shown in Fig. \ref{fig:hep_2_hpf_ce} is more clearly characterized by discrete speckles distributed throughout the interphase nuclei and characteristically found in the condensed nuclear chromatin in comparison with that in Fig. \ref{fig:hep_2_ce}. The coarse speckled cell image as shown in Fig. \ref{fig:hep_2_hpf_cs} reveals more distinctly a coarse granular nuclear staining of the interphase cell in comparison with that in Fig. \ref{fig:hep_2_cs}. The cytoplasmic cell image as shown in Fig. \ref{fig:hep_2_hpf_cy} is more sharply represented by fine fluorescent fibers in comparison with that in Fig. \ref{fig:hep_2_cy}. The fine speckled cell image as shown in Fig. \ref{fig:hep_2_hpf_fs} in comparison with that in Fig. \ref{fig:hep_2_fs} makes the visualization of the fine granular nuclear staining of the interphase cell nuclei clearer. The homogeneous cell image as shown in Fig. \ref{fig:hep_2_hpf_ho} shows only distinctive solid staining of the entire nucleus of interphase cells in comparison with that in Fig. \ref{fig:hep_2_ho}. The nucleolar cell image (Fig. \ref{fig:hep_2_hpf_nu}) is represented by more clearly clustered large granules in the nucleoli of interphase cells in comparison with that in Fig. \ref{fig:hep_2_nu}. In particular, Fig. \ref{fig:hep_2_hpf_fs}, Fig. \ref{fig:hep_2_hpf_ho}, and Fig. \ref{fig:hep_2_hpf_nu} are more distinctive prior to the filtering of each cell image.\\

\noindent {\bf Feulgen-stained cells}: Buyssens {\it{et al.}} \cite{buyssens2013multiscale} studied the classification of Feulgen stained cells. The dataset $\footnote{https://sites.google.com/site/pierrebuyssens/cells-database}$ is available for academic purposes. Feulgen-stained cells \cite{ten1993method, nielsen2012automatic} are composed of six different classes with 215 abnormal mesothelials, 209 dystrophic mesothelials, 201 normal mesothelials, 195 macrophages, 198 polynuclears, and 196 lymphocytes, as shown in Fig. \ref{fig:fsc-noLoG}. The performances of the classification are measured by using a 10-fold cross-validation with 3 RGB channels.

% ========================= figure 10 ===================================
\begin{figure}[b]
	\centering
	\subfigure[Abnormal mesothelials.]{
		\label{fig:fsc-LoG-AM}
		\includegraphics[width=0.29\linewidth]{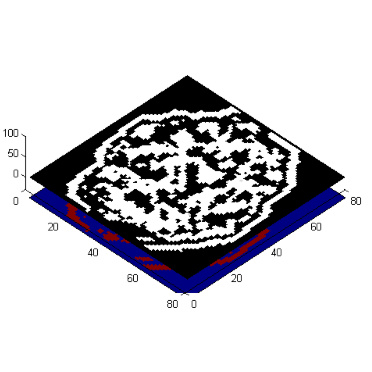}
	}
	\subfigure[Dystrophic mesothelials.]{
		\label{fig:fsc-LoG-DM}
		\includegraphics[width=0.29\linewidth]{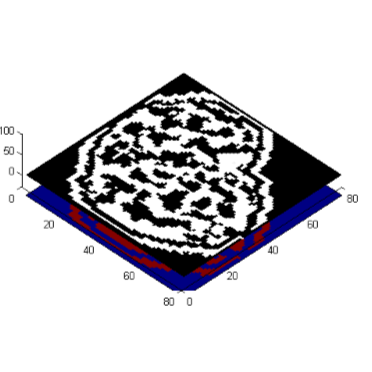}
	}
	\subfigure[Lymphocyte.]{
		\label{fig:fsc-LoG-Ly}
		\includegraphics[width=0.29\linewidth]{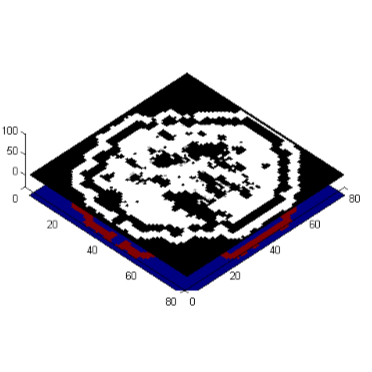}
	}\hfill
	\subfigure[Macrophages.]{
		\label{fig:fsc-LoG-Ma}
		\includegraphics[width=0.29\linewidth]{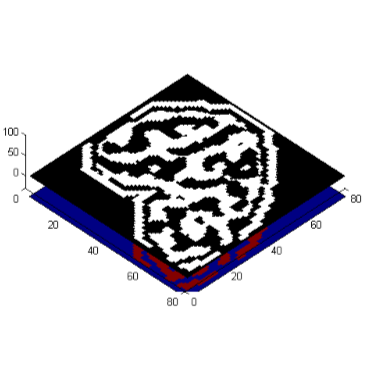}
	}
	\subfigure[Normal mesothelials]{
		\label{fig:fsc-LoG-NM}
		\includegraphics[width=0.29\linewidth]{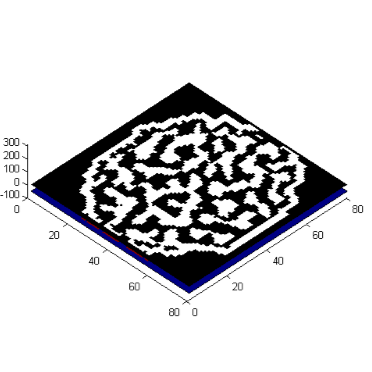}
	}
	\subfigure[Polynuclear.]{
		\label{fig:fsc-LoG-Po}
		\includegraphics[width=0.29\linewidth]{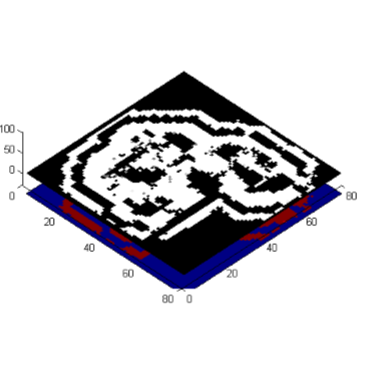}
	}
	\caption{Examples of Feulgen stained cell images with LOG filter shown in 3D of surf, Matlab.}
	\label{fig:fsc-LoG}
\end{figure} 

Feulgen-stained cell images are filtered by the Laplacian of Gaussian (LOG) filter \cite[pp. 581-585]{gonzalez2002woods} so that discriminative features are induced by their texture information. As the LOG filter $H(u,v)$ is derived by the 2D Gaussian distribution, it reduces noise, smooths the image, and establishes the location of edges and high frequencies. The detailed structures of Feulgen-stained cell images can be extracted by optimal sigma $\sigma = 0.2$ and a $5\times5$ mask. The output of LOG filtering in Fig. \ref{fig:feat_ex}: filtered image is given by $G(u,v) = F(u,v)H(u,v)$, where $F(u,v)$ are the Feulgen stained cell images shown in Fig. \ref{fig:fsc-noLoG}. The LOG filtering images $g(x,y)$ are converted by inverse DFT. Fig. \ref{fig:fsc-LoG} shows the results of the LOG filtering of the cell images visualized in 3D using grayscale images. The filtered cell images reveal their edge structures as components of high frequency that can be used for discriminative features. The lymphocyte cell image as shown in Fig. \ref{fig:fsc-LoG-Ly}, the macrophage cell image as shown in Fig. \ref{fig:fsc-LoG-Ma}, and the polynuclear cell image as shown in Fig. \ref{fig:fsc-LoG-Po} fall under more distinct classes due to their low frequencies and contours of the cell images in comparison with others, as shown in Fig. \ref{fig:fsc-nonLoG-Ly}, Fig. \ref{fig:fsc-nonLoG-Ma}, and  Fig. \ref{fig:fsc-nonLoG-Po}. The abnormal mesothelial cell image (Fig. \ref{fig:fsc-LoG-AM}), the dystrophic mesothelial cell image (Fig. \ref{fig:fsc-LoG-DM}), and the normal mesothelial cell image (Fig. \ref{fig:fsc-LoG-NM}) are more clearly depicted with the components of high frequencies in comparison with others, as shown in Fig. \ref{fig:fsc-nonLoG-AM}, Fig. \ref{fig:fsc-nonLoG-DM}, and Fig. \ref{fig:fsc-nonLoG-NM}. 

% ========================= figure 11 ===================================
\begin{figure} [h]
	\centering
	\subfigure[The SPNs.]{
		\label{fig:architecture_dspn}
		\includegraphics[width=0.6\linewidth]{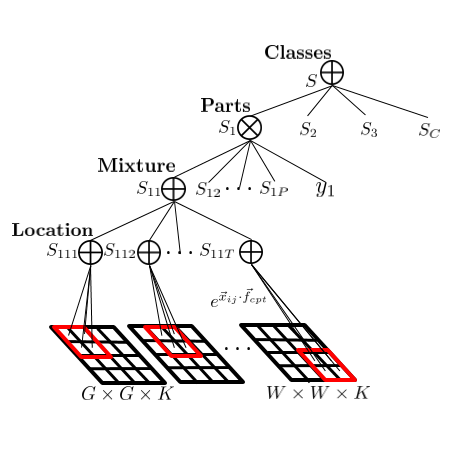}}
	\subfigure[The t-SPNs of sub-SPNs.]{
		\label{fig:architecture_sub}
		\includegraphics[width=0.82\linewidth]{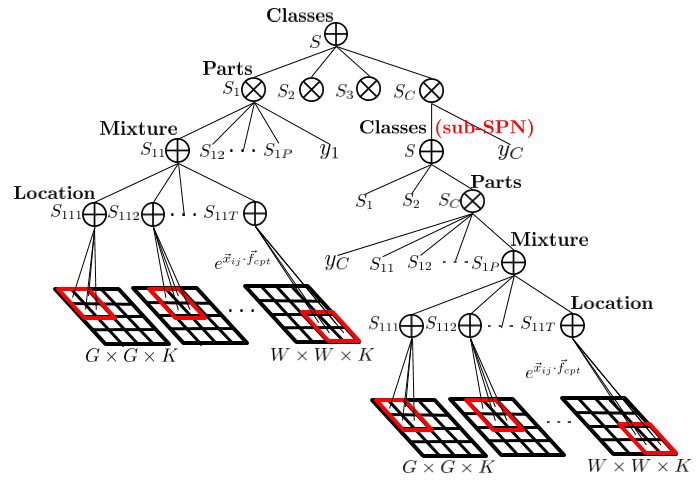}
	}
	\caption{Two kinds of SPNs's architectures for cell classification.}	
	\label{fig:architecture}
\end{figure} 

\subsection{Experimental setup} \label{sec:exp_setup}
%SPNs and Sub-SPNs
Based on the features (${8}\times{8}\times{1600}$) in Section \ref{sec:dis_feat_ext}, both SPNs and t-SPNs can be learned from the architectures as shown in Fig. \ref{fig:architecture}. The architectures of two kinds of SPNs consist of each root (sum) node for soft inference, product nodes for the classes, and the rest (max nodes) for hard inference. One of the architectures as shown in Fig. \ref{fig:architecture_dspn} follows the basic structure \cite{gens2012discriminative} used for CLL-based discriminative learning of SPNs. In our experiments, the architectures ($S$) are also used in MM-based learning of SPNs and are made up of C classes, P parts ($S_c$) per class, T mixture components ($S_{cp}$) per part and location ($S_{cpt}$) summing over the positions ($i,j$) of the feature matrix ($G\times{G}\times{K}$) with weights ($\vec{f}_{cpt}$). The parameters set $C = 6, P = 10$, and $T = 25$ in all experiments. The parameters of another architecture as shown in Fig. \ref{fig:architecture_sub} for the t-SPNs have the same values as those of the SPNs except for the number of classes ($C$) that depends on the characteristics of each dataset. The HEp-2 dataset has three confusing classes: find speckled, homogeneous, and nucleolar, so that $C = 3$. The Feulgen dataset has two confusing classes: dystrophic mesothelials and normal mesothelials, so that $C = 2$ based on the confusion matrices processed by the first architecture (Fig. \ref{fig:architecture_dspn}). In addition, the L2-regularization (REG) of the objective function is evaluated in Eq. \ref{eq:obj_fcn}. The evaluation of this regularization term is based on both the HEp-2 dataset with HPF and the Feulgen dataset with LOG filtering. Each REG parameter depends on its dataset; the HEp-2 dataset is set by $\beta = 0.015$ and the Feulgen stained dataset is set by $\beta = 0.010$.

\subsection{Experimental results of the two datasets}\label{sec:exp_results} % ch. 5.2.

To demonstrate the effectiveness of the proposed t-SPNs+MM and the filterings: HPF and LOG filtering, the following four methods as shown in Table. \ref{table:SPNs_performance} are compared with each other on two benchmark datasets: HEp-2 dataset and Feulgen stained dataset. Table. \ref{table:SPNs_performance} presents the quantitative results of the four baseline experiments with two filtering methods. The results of the experiments with HPF are superior to those without HPF on the HEp-2 dataset. The same result is found for the experiments with LOG filter on the Feulgen stained dataset. Both HPF and LOG filters help the feature extractor build more discriminative dictionary features to improve the classification results. As shown above, the filtering for each dataset plays a decisive role in classifying cell images into each class. 

% ========================= figure 12 ===================================
\begin{figure}
	\centering
	\subfigure[SPNs w/ HPF.]{
		\label{fig:hep2_spns}
		\includegraphics[width=0.46\linewidth]{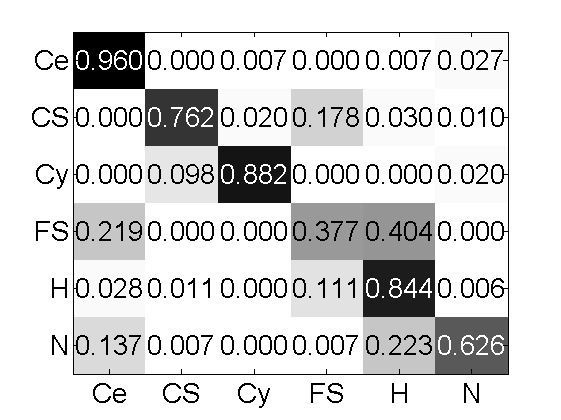}
	}
	\subfigure[SPNs+MM w/ HPF.]{
		\label{fig:hep2_spns_mm}
		\includegraphics[width=0.46\linewidth]{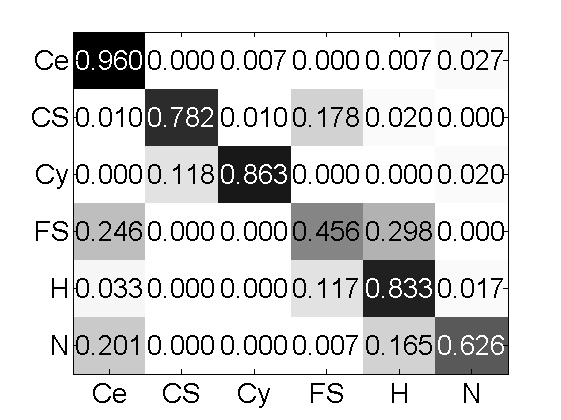}
	} \hfill
	\subfigure[t-SPNs w/ HPF.]{
		\label{fig:hep2_subspns}
		\includegraphics[width=0.46\linewidth]{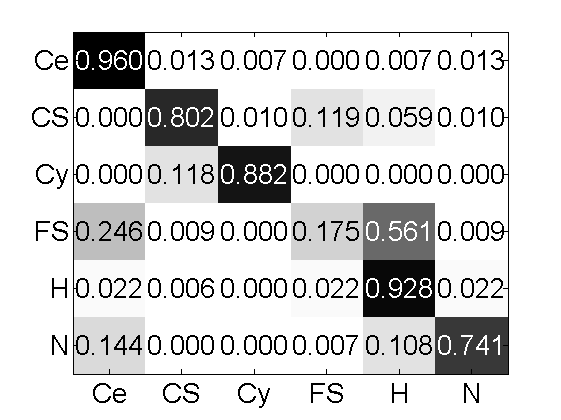}
	}
	\subfigure[t-SPNs+MM w/ HPF.]{
		\label{fig:hep2_subspns_mm}
		\includegraphics[width=0.46\linewidth]{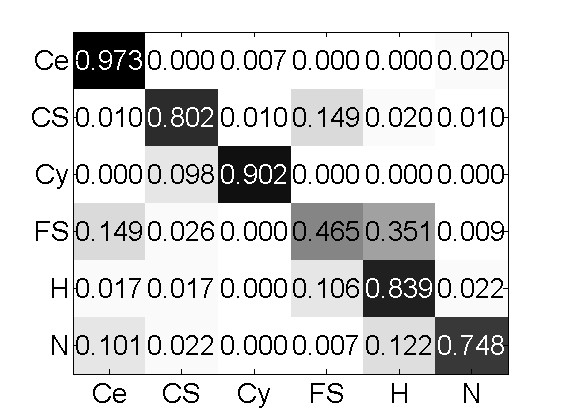}
	}
	\caption{Effectiveness of the t-SPNs+MM with HPF on HEp-2 dataset where Ce:Centromere, CS:Coarse speckled, Cy:Cytoplasmatic, FS:Fine speckled, H:Homogeneous, and N:Nucleolar.}
	\label{fig:hep2_com_spn}
\end{figure}

\begin{table}% table 1
	\begin{center}
		\caption{Classification accuracy comparison with four different methods on two benchmark datasets, HEp-2 cells and Feulgen stained cells with 10-fold cross-validation.}
		\label{table:SPNs_performance}
		\begin{tabular}{c|c|c|c|c}
			\hline
			\multicolumn{1}{c|}{\multirow{3}{*}{Method} }  & \multicolumn{4}{|c}{Datasets} \\
			\cline{2-5}
			\multicolumn{1}{c|}{} & \multicolumn{2}{|c|}{HEp-2 cells} & \multicolumn{2}{|c}{Feulgen stained cells}  \\
			\cline{2-5}
			\multicolumn{1}{c|}{} & w/o HPF & {\bf w/ HPF} & w/o LOG & {\bf w/ LOG} \\
			\hline\hline
			SPNs              & 71.4\% & 74.5\%      & 90.1\% & 90.9\%  \\
			SPNs+MM           & 74.3\% & 75.6\%      & 90.5\% & 91.5\%  \\
			t-SPNs          & 72.8\% & 76.2\%      & 90.7\% & 91.1\%  \\
			{\bf t-SPNs+MM} & 75.3\% & {\bf 79.0\%}& 92.1\% & {\bf 92.7\% }\\
			\hline
		\end{tabular}
	\end{center}
\end{table}

% the results of sub-SPNs+MM, effectiveness of sub-SPNs+MM
Without filtering, the performance of SPNs+MM among the four baseline experiments is better than that of the SPNs on two datasets. The combination of the t-SPNs and MM has the best performance among the four baseline experiments. This result shows that the sub-SPNs of the t-SPNs regard confusing classes as one class, and they are advantageous to the joint probabilistic evaluation. The MM-based learning method improves the classification results by reducing the generalization errors of the confusing classes. 

\begin{figure}
	\centering
	\subfigure[SPNs w/ LOG.]{
		\label{fig:fsc_spns}
		\includegraphics[width=0.46\linewidth]{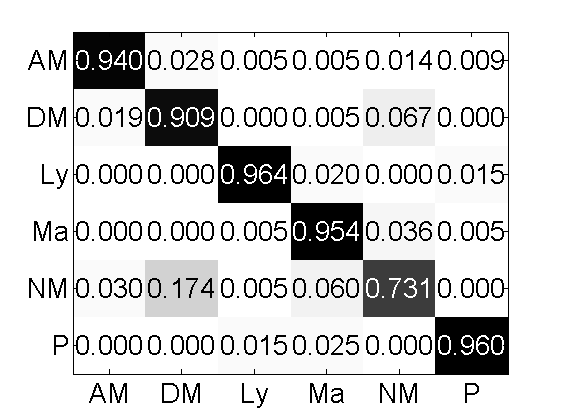}
	}
	\subfigure[SPNs+MM w/ LOG.]{
		\label{fig:fsc_spns_mm}
		\includegraphics[width=0.46\linewidth]{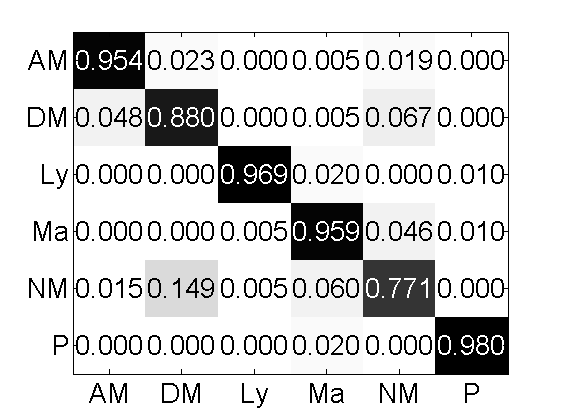}
	} \hfill
	\subfigure[t-SPNs w/ LOG.]{
		\label{fig:fsc_subspns}
		\includegraphics[width=0.46\linewidth]{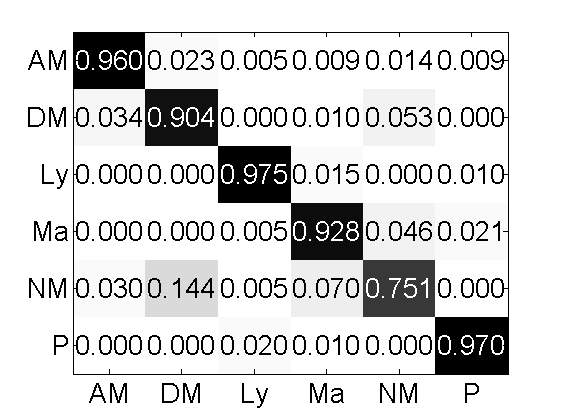}
	}
	\subfigure[t-SPNs+MM w/ LOG.]{
		\label{fig:fsc_subspns_mm}
		\includegraphics[width=0.46\linewidth]{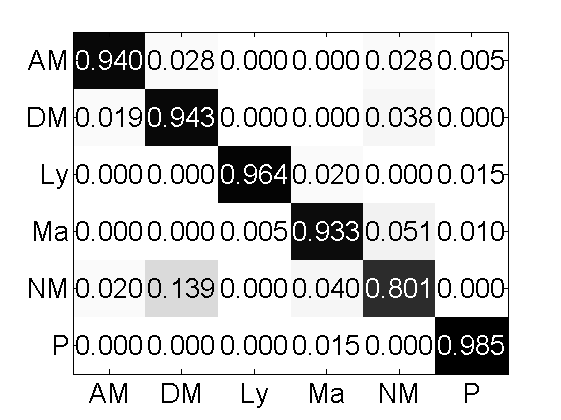}
	}
	\caption{Effectiveness of the t-SPNs+MM with LOG filter on Feulgen stained dataset where AM:Abnormal mesothelials, DM:Dystrophic mesothelials, Ly:Lymphocyte, Ma:Macrophages, NM:Normal mesothelials, and P:Polynuclear.}
	\label{fig:fsc_com_spn}
\end{figure}

The confusion matrices of the four baseline experiments show the detailed analysis of the effectiveness of t-SPNs+MM with each filter in Fig. \ref{fig:hep2_com_spn} and Fig. \ref{fig:fsc_com_spn}. As shown in Fig. \ref{fig:hep2_spns} and Fig. \ref{fig:fsc_spns}, fine-speckled (FS), homogeneous (H), and nucleolar (N) cells are the most confusing classes in the HEp-2 cells, and dystrophic mesothelials (DM) and normal mesothelials (NM) are the most confusing classes in the Feulgen stained cells because their recognition (true positive) rate is very low. In Fig. \ref{fig:hep2_com_spn} on the HEp-2 dataset, the performances of CS and FS are slightly improved by SPNs+MM /w HPF (Fig. \ref{fig:hep2_spns_mm}) compared with those of the classes estimated by the SPNs /w HPF (Fig. \ref{fig:hep2_spns}). The t-SPNs /w HPF (Fig. \ref{fig:hep2_subspns}) further enhances the accuracy of CS and the confusing classes for H and N. The t-SPNs+MM /w HPF (Fig. \ref{fig:hep2_subspns_mm}) shows the overall improvements of the HEp-2 cell classification. In Fig. \ref{fig:fsc_com_spn} on the Feulgen stained dataset, the result of NM is improved by SPNs+MM w/ LOG (Fig. \ref{fig:fsc_spns_mm}), although the accuracy of DM is worse than that of NM in Fig. \ref{fig:fsc_spns}. The sub-SPNs (Fig. \ref{fig:fsc_subspns}) of t-SPNs reduces the confusedness between DM and NM. The t-SPNs+MM /w LOG (Fig. \ref{fig:fsc_subspns_mm}) enhances the performance of Feulgen stained cell classification and reduces the ambiguity of confusing classes for DM and NM. As a result, the combined t-SPNs+MM with filtering can effectively focus on confusing cell classification through the generality of MM-based learning and the joint probabilistic evaluation of t-SPNs. 

% ========================= figure 13 ===================================

\subsection{Results of the regularization}\label{sec:exp_reg}
The effectiveness of the REG term of the hinge loss functions in Eq. \ref{eq:obj_fcn} is proved by the four baseline experiments on both the HEp-2 dataset with HPF (where the regularization parameter $\beta$ is set to $0.015$) and the Feulgen stained dataset with LOG filter ($\beta = 0.010$). Table. \ref{table:SPNs_reg_performance} shows the results of the evaluation. The performances of all four experiments with the regularization term surpass those without it. The results with respect to the HEp-2 dataset increase the performances by approximately 1\% on average. The accuracy $79.6 \%$ of t-SPNs+MM+REG /w HPF is better than any other methods on the HEp-2 dataset. The results with respect to the Feulgen stained dataset slightly improve the performances. The best result $92.8 \%$ is also performed by t-SPNs+MM+REG /w LOG on the Feulgen stained dataset. 
\begin{table} % table 2
	\begin{center}
		\caption{Effectiveness of objective function of the regularization term with four different methods on HEp-2 dataset.}
		\label{table:SPNs_reg_performance}
		\begin{tabular}{c|c|c|c|c}
			\hline
			\multicolumn{1}{c|}{\multirow{3}{*}{Method} }  & \multicolumn{4}{|c}{Datasets} \\
			\cline{2-5}
			\multicolumn{1}{c|}{} & \multicolumn{2}{|c|}{HEp-2 cells} & \multicolumn{2}{|c}{Feulgen stained cells}  \\
			\cline{2-5}
			\multicolumn{1}{c|}{} & w/HFP & {\bf REG w/ HPF} & w/LOG& {\bf REG w/ LOG}\\
			\hline\hline
			SPNs            & 74.5\% & 75.4\%      & 90.9\% & 91.5\% \\
			SPNs+MM         & 75.6\% & 76.1\%      & 91.5\% & 91.6\% \\
			t-SPNs          & 76.2\% & 77.9\%      & 91.1\% & 91.8\% \\
			{\bf t-SPNs+MM} & 79.0\% & {\bf 79.6\%}& 92.7\% & {\bf92.8\%} \\
			\hline
		\end{tabular}
	\end{center}
\end{table}

\begin{figure} [t]
	\centering
	\subfigure[SPNs+REG w/ HPF.]{
		\label{fig:hep2_spns_reg}
		\includegraphics[width=0.46\linewidth]{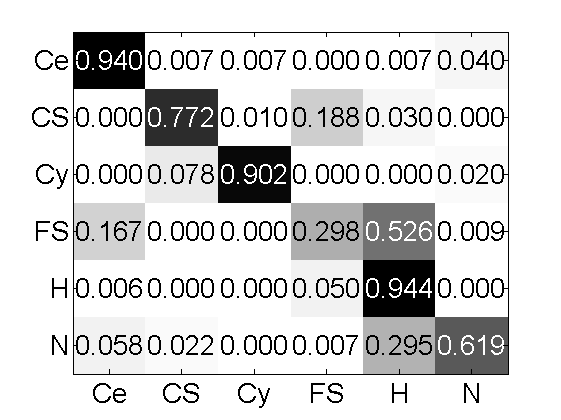}
	}
	\subfigure[SPNs+MM+REG w/ HPF.]{
		\label{fig:hep2_spns_mm_reg}
		\includegraphics[width=0.46\linewidth]{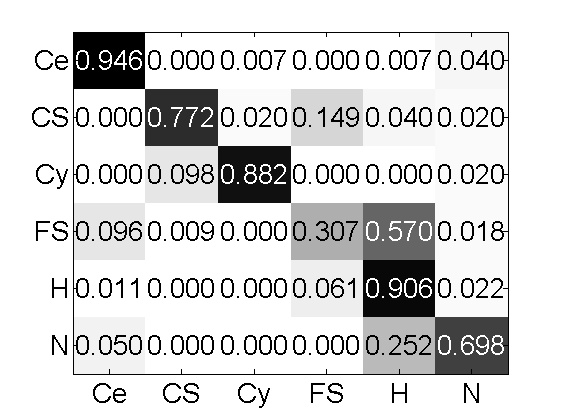}
	} \hfill
	\subfigure[t-SPNs+REG w/ HPF.]{
		\label{fig:hep2_subspns_reg}
		\includegraphics[width=0.46\linewidth]{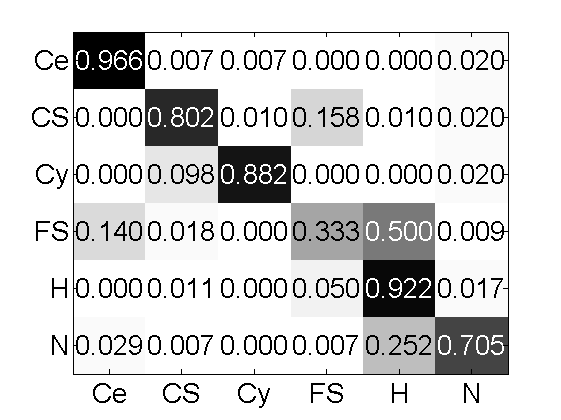}
	}
	\subfigure[t-SPNs+MM+REG w/ HPF.]{
		\label{fig:hep2_subspns_mm_reg}
		\includegraphics[width=0.46\linewidth]{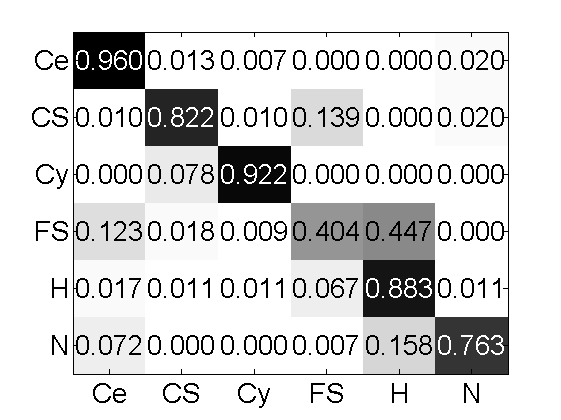}
	}
	\caption{Effectiveness of the t-SPNs+MM+REG with HPF on HEp-2 dataset where Ce:Centromere, CS:Coarse speckled, Cy:Cytoplasmatic, FS:Fine speckled, H:Homogeneous, and N:Nucleolar.}
	\label{fig:hep2_com_spn_reg}
\end{figure}

\begin{figure} [t]
	\centering
	\subfigure[SPNs+REG w/ LOG.]{
		\label{fig:fsc_spns_reg}
		\includegraphics[width=0.46\linewidth]{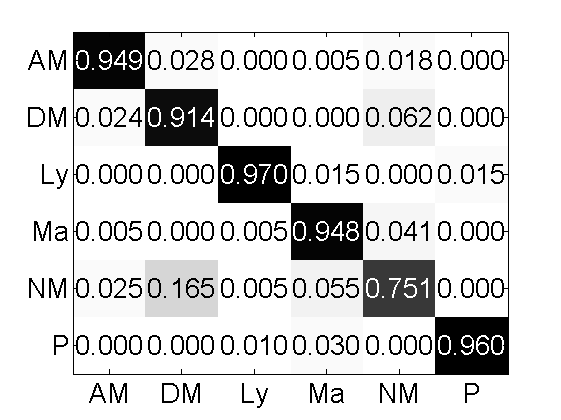}
	}
	\subfigure[SPNs+MM+REG w/ LOG.]{
		\label{fig:fsc_spns_mm_reg}
		\includegraphics[width=0.46\linewidth]{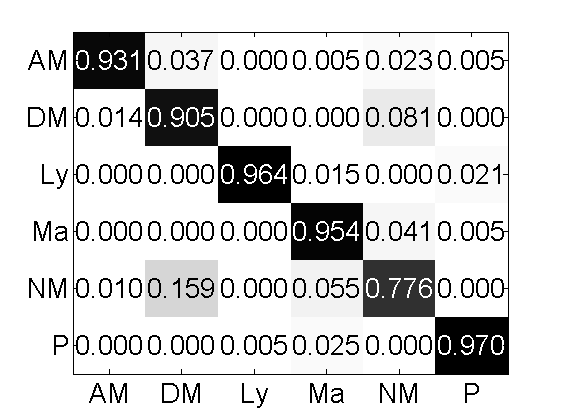}
	} \hfill
	\subfigure[t-SPNs+REG w/ LOG.]{
		\label{fig:fsc_subspns_reg}
		\includegraphics[width=0.46\linewidth]{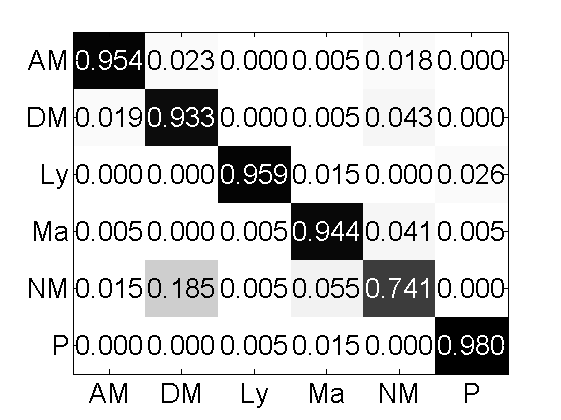}
	}
	\subfigure[t-SPNs+MM+REG w/ LOG.]{
		\label{fig:fsc_subspns_mm_reg}
		\includegraphics[width=0.46\linewidth]{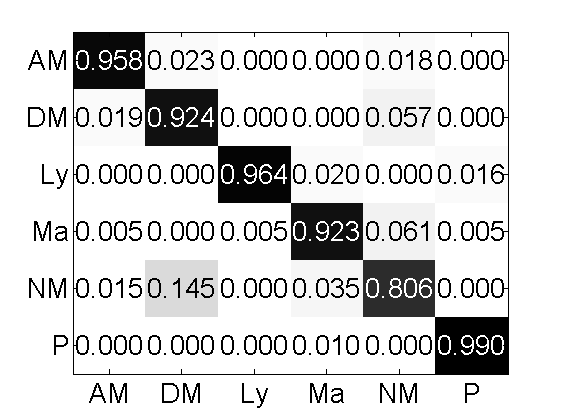}
	}
	\caption{Effectiveness of the t-SPNs+MM+REG with LOG filter on Feulgen stained dataset where AM:Abnormal mesothelials, DM:Dystrophic mesothelials, Ly:Lymphocyte, Ma:Macrophages, NM:Normal mesothelials, and P:Polynuclear.}
	\label{fig:fsc_com_spn_reg}
\end{figure}

The further generalization of cell classification through REG can be confirmed by the confusion matrices. Fig. \ref{fig:hep2_com_spn_reg} shows the confusion matrices of the results with respect to REG. The improvements in the accuracy can be observed by comparing the confusion matrices with REG as shown in Fig. \ref{fig:hep2_com_spn_reg} and those without REG as shown in Fig. \ref{fig:hep2_com_spn}. Clearly, t-SPNs+MM+REG /w HPF decreases the overall misclassification rates among the FS, H, and N classes on the HEp-2 dataset, as shown in Fig. \ref{fig:hep2_subspns_reg} and Fig. \ref{fig:hep2_subspns_mm_reg}. Fig. \ref{fig:fsc_com_spn_reg} shows the confusion matrices of the results on the Feulgen stained dataset. SPNs induce the most enhancement with REG. In Fig.\ref{fig:fsc_spns_reg}, the classes AM, DM, Ly, and NM are more generalized than those in Fig. \ref{fig:fsc_spns}. In Fig. \ref{fig:fsc_spns_mm_reg}, Fig. \ref{fig:fsc_subspns_reg}, and Fig. \ref{fig:fsc_subspns_mm_reg}, the improvements are small in comparison with those in Fig. \ref{fig:fsc_com_spn}.

% ========================= figure 14 ===================================
% ========================= figure 15 ===================================
% ========================= figure 16 ===================================
\begin{figure}
	\centering
	\subfigure[Fused (SRC) proposed by Theodorakopoulos {\it et al.} \cite{theodorakopoulos2014hep}.]{
		\label{fig:hep2_src}
		\includegraphics[width=0.46\linewidth]{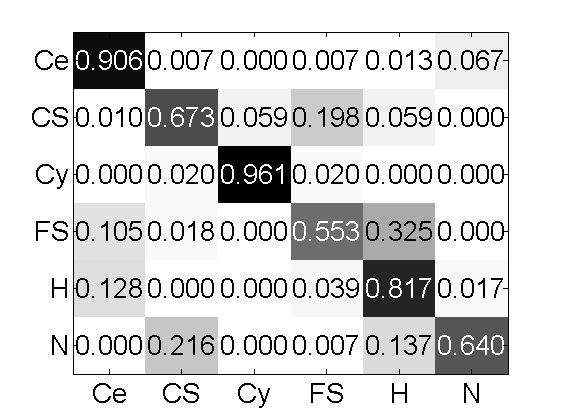}
	}
	\subfigure[Human expert \cite{ensafi2014automatic}.]{
		\label{fig:hep2_human}
		\includegraphics[width=0.46\linewidth]{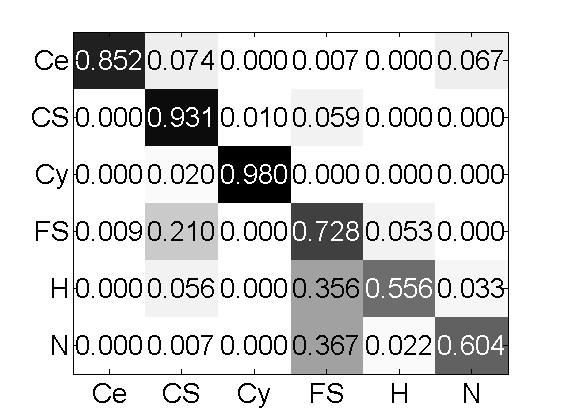}
	}
	\caption{The performances of the fused (SRC) and human expert on HEp-2 dataset where Ce:Centromere, CS:Coarse speckled, Cy:Cytoplasmatic, FS:Fine speckled, H:Homogeneous, and N:Nucleolar.}
	\label{fig:hep2_com}
\end{figure}

\begin{table} % table 3
	\begin{center}
		\caption{Comparison of average test accuracy on ten-folds of Feulgen stained dataset.}
		\label{table:Comparison_Feulgen}
		\begin{tabular}{cc}
			\hline
			Method                                   & Accuracy ($\pm{\sigma}$)\\
			\hline
			CNN$_{28}$ \cite{buyssens2013multiscale} & 90.96\% ($\pm$1.07\%)\\
			CNN$_{40}$ \cite{buyssens2013multiscale} & 91.98\% ($\pm$1.02\%)\\
			CNN$_{56}$ \cite{buyssens2013multiscale} & 91.99\% ($\pm$0.98\%)\\
			CNN$_{80}$ \cite{buyssens2013multiscale} & 92.45\% ($\pm$0.89\%)\\
			{\bf{t-SPNs+MM+REG w/ LOG (Ours)}}       & {\bf 92.75\%($\pm$3.69\%)}\\
			MCNN\cite{buyssens2013multiscale}        & 93.98\% ($\pm$0.82\%)\\
			wMCNN \//w CNN$_{28}$ \cite{buyssens2013multiscale} & 94.26\% ($\pm$0.79\%)\\
			\hline
		\end{tabular}
	\end{center}
\end{table}

\begin{table} % table 4
	\begin{center}
		\caption{Comparison of test accuracy on HEp-2 dataset.}
		\label{table:Comparison_HEp-2}
		\begin{tabular}{cc}			
			\hline
			Method                                                 & Accuracy\\
			\hline
			Xiangfei {\it et al.}\cite{kong2014hep}                & 67.0\% \\
			Nosaka {\it et al.}\cite{nosaka2014hep}                & 68.5\% \\
			Boesen Lindbo Larsen {\it et al.}\cite{larsen2014hep}  & 71.5\% \\ 
			Ensafi {\it et al.}\cite{ensafi2014automatic}          & 72.8\%	\\
			Human expert \cite{ensafi2014automatic}                & 73.3\% \\
			Shen {\it et al.} \cite{shen2014hep}                   & 74.4\% \\
			Theodorakopoulos {\it et al.} \cite{theodorakopoulos2014hep} & 75.1\% \\
			{\bf{Ours (t-SPNs+MM+REG w/ HPF)}}                     & {\bf 79.6\%} \\
			\hline
		\end{tabular}
	\end{center}
\end{table}

\subsection{Comparison with other algorithms}\label{sec:exp_com}
The performances of t-SPN-based algorithms are compared with those of other algorithms. Table. \ref{table:Comparison_HEp-2} shows the comparison of test accuracy on the HEp-2 dataset with that in other algorithms. The proposed algorithm outperforms the previous state-of-the-art result performed by Theodorakopoulos {\it et al.}\cite{theodorakopoulos2014hep}, human expert \cite{ensafi2014automatic} as well as all participants: Xiangfei {\it et al.}\cite{kong2014hep}, Nosaka {\it et al.}\cite{nosaka2014hep} of the \textit{HEp-2 Cell Classification Contest} \cite{foggia2013benchmarking}. Fig. \ref{fig:hep2_com} shows the confusion matrices for the performances of the fused (SRC) proposed by Theodorakopoulos {\it et al.} \cite{theodorakopoulos2014hep} and the human expert \cite{ensafi2014automatic}. Comparing Fig. \ref{fig:hep2_subspns_mm} and Fig. \ref{fig:hep2_src} and Fig. \ref{fig:hep2_human}, the accuracy of t-SPNs+MM+REG w/ HPF outperforms that of the fused (SRC) except for CS and Cy, and it is better than that of the human expert in Ce, H, and N classes. This finding proves that t-SPNs+MM+REG w/ HPF is effective for confusing classes. With regard to the Feulgen stained dataset, Table. \ref{table:Comparison_Feulgen} compares our results with those of CNN models proposed by Buyssens {\it el al.} \cite{buyssens2013multiscale}. The performance of t-SPNs+MM+REG w/ LOG is better than that of CNN models with single-scaled input images, except for both multi-scale CNN (MCNN) and wMCNN \//w CNN$_{28}$. The basic assumption of MCNN is that the lowest resolution may be less salient than the full resolution. With this assumption, the error rates for the CNNs decrease while the resolution of the inputs increase in the experiments, as the CNNs are able to capture more information. Then, the outputs of the MCNN, the best approach, are then computed as a linear combination of the outputs of the N CNNs. However, our feature extraction method performs with the max pooling process. That is, the discriminative is rarely influenced by the size of the images. Nevertheless, the multi-scale inputs can indeed be one of the possible solutions to cell classification tasks.

% -------------------------------------------------------------------------------------
%                      Chapter 6. Conclusion 
% -------------------------------------------------------------------------------------
\section{Conclusion}\label{sec:conc}
In this study, the architecture referred to as a tree-structured sum-product network (t-SPN) is proposed for cell classification. The t-SPN is built such that the unnormalized probability is represented as conditional probabilities of a subset of most similar cell classes. The constructed t-SPN architecture was learned by maximizing the margin, defined as the difference in the conditional probability between the true and the most competitive false label. To enhance the generalization ability of the architecture, L2-regularization (REG) was considered along with the maximum margin (MM) criterion in the learning process. Furthermore, to highlight cell features, this paper investigated the effectiveness of two generic high-pass filters: ideal high-pass filtering and the Laplacian of Gaussian (LOG) filtering. On both HEp-2 and Feulgen benchmark datasets, the t-SPN architecture learned based on the max-margin criterion with regularization produced the highest accuracy rate compared to other state-of-the-art algorithms that include convolutional neural network (CNN) based algorithms. In addition, the experimental results indicate that the ideal high-pass filter is more effective on the HEp-2 dataset, which is based on immunofluorescence staining, while the LOG is more effective on Feulgen dataset, which is based on Feulgen staining.

\ifCLASSOPTIONcaptionsoff
  \newpage
\fi

\bibliographystyle{plain}
\bibliography{ieee_stsp}

\vspace{-10mm}

\begin{IEEEbiography}[{\includegraphics[width=1in,height=1.25in,clip,keepaspectratio]{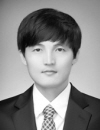}}]{Haeyong Kang}
(S'05) received the M.S. degree in systems and information engineering from University of Tsukuba in 2007. From April 2007 to October 2010, he worked as an associate research engineer at LG Electronics. With working experiences at the Korea Institute of Science and Technology (KIST) and the University of Tokyo, He is currently pursuing the Ph.D at School of Electrical Engineering, KAIST. His current research interests include machine learning for object classification, multimedia signal processing, and discriminative training.
\end{IEEEbiography}

\vspace{-10mm}

\begin{IEEEbiography}[{\includegraphics[width=1in,height=1.25in,clip,keepaspectratio]{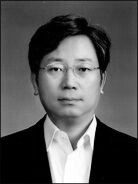}}]{Chang D. Yoo}
	Chang D. Yoo (S’92-M’96-SM’11) received the	B.S. degree in Engineering and Applied Science
	from the California Institute of Technology, the M.S. degree in Electrical Engineering from Cornell University and the Ph.D degree in Electrical Engineering from the Massachusetts Institute of Technology. From January 1997 to March 1999 he was Senior Researcher at Korea Telecom (KT). Since 1999, he has been on the faculty in the School of Electrical
	Engineering at the Korea Advance Institute of Science and Technology (KAIST). He also served as
	Dean of the Office of Special Projects (2011-2013) and Dean of the Office of International Relations (2013-2015). He has been a consultant for many firms and is currently a consultant for the Korean Foundation for Advance Studies (one of the oldest and largest scholarship foundations in Korea). He is also Director of Korea Institute of Electrical Engineers (KIEE) and Director of the Acoustical Society of Korea (ASK). He is Member of Tau Beta Pi and Sigma Xi. He was on the technical committee member of IEEE machine learning for signal processing society from 2009 to 2011. He had also served as Associated Editor of IEEE Signal Processing Letters (2011-2012), IEEE Transactions on Information Forensics and Security (2012-2013), IEEE Transactions on Audio,
	Speech and Language Processing (2011-2014).
\end{IEEEbiography}

\vspace{-10mm}

\begin{IEEEbiography}[{\includegraphics[width=1in,height=1.25in,clip,keepaspectratio]{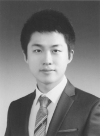}}]{Yongcheon Na}
(S'13-M'15) received the B.S. degree in electronics and computer engineering from the Hanyang University in 2013, and the M.S. degree in Interdisciplinary Program for Future Vehicles from the Korea Advanced Institute of Science and Technology (KAIST) in 2015. His research interests include machine learning for object classification, multimedia signal processing, and discriminative training.
\end{IEEEbiography}

%\begin{IEEEbiographynophoto}
\end{document}